# CongNaMul: A Dataset for Advanced Image Processing of Soybean Sprouts

Byunghyun Ban*, Donghun Ryu, Su-won Hwang
Department of Research
Imagination Garden Inc.
Andong-si, Republic of Korea
{halfbottle*, dhryu, hsw}@sangsang.farm

*Abstract*—We present 'CongNaMul', a comprehensive dataset designed for various tasks in soybean sprouts image analysis. The CongNaMul dataset is curated to facilitate tasks such as image classification, semantic segmentation, decomposition, and measurement of length and weight. The classification task provides four classes to determine the quality of soybean sprouts: normal, broken, spotted, and broken and spotted, for the development of AI-aided automatic quality inspection technology. For semantic segmentation, images with varying complexity, from single sprout images to images with multiple sprouts, along with human-labelled mask images, are included. The label has 4 different classes: background, head, body, tail. The dataset also provides images and masks for the image decomposition task, including two separate sprout images and their combined form. Lastly, 5 physical features of sprouts (head length, body length, body thickness, tail length, weight) are provided for image-based measurement tasks. This dataset is expected to be a valuable resource for a wide range of research and applications in the advanced analysis of images of soybean sprouts. Also, we hope that this dataset can assist researchers studying classification, semantic segmentation, decomposition, and physical feature measurement in other industrial fields, in evaluating their models. The dataset is available at the authors' repository(https://bhban.kr/data).

## I. INTRODUCTION

### A. Soybean Sprout Industry

Soybean sprouts are a unique type of vegetable cultivated from soybeans, characterized by their short-term cultivation in darkness. Plants undergo photosynthesis when they receive light in their sprouting stage, increasing the density of chlorophyll and forming green leaves. In contrast, soybean sprouts must be thoroughly shielded from light during cultivation to prevent photosynthesis. By placing yellow soybeans in a dark room and periodically sprinkling enough water, white roots sprout and gradually lengthen. However, since they do not undergo photosynthesis and do not form chlorophyll, the head color does not turn green, and only the length of the body continues to grow. After about 7 days of cultivation, soybean sprouts of about 8~15cm in length are produced. This is similar to the method of producing mung bean sprouts.

Unlike mung bean sprouts, soybean sprouts are mainly consumed in Korea and are used as ingredients in various Korean dishes such as soup, salad, and bulgogi. The size of the soybean sprout industry in Korea is estimated to be around 300 billion won, and it is estimated that each Korean consumes more than 10kg of soybean sprouts per year [1].

### B. Massproduction of Soybean Sprout

The traditional cultivation method of soybean sprouts proceeds as follows. First, the soybeans are washed and soaked in water for more than three days. The soaked beans are then placed in a ceramic jar known as a "Siru" and covered with water. The bottom of the Siru is perforated with numerous small holes, allowing water to drain easily and preventing the beans from rotting. The Siru is equipped with a lid that prevents the beans from receiving direct light. In this state, water is added daily, and the beans are cultivated for more than seven days, which is the traditional method of producing soybean sprouts.

Due to the high consumption and preference for soybean sprouts among Koreans, various processes for mass production of soybean sprouts have been researched for decades. Some factories use large rubber bowls as a substitute for the Siru, while others use massive stainless steel tanks. In large-scale facilities, the key to the automated production process of soybean sprouts is to periodically add water using automated machines while blocking light.

### C. Differences from Plant Factory Systems

The term "plant factory" is often used interchangeably with "vertical farm". The core concept of a plant factory is to artificially create an environment similar to nature, enabling optimal growth conditions for plants. For instance, using LEDs to provide artificial light allows plants to photosynthesize as if they were outdoors, and precisely controlling air factors such as temperature, humidity, and $CO_2$ concentration enables plants to grow well. This is the core philosophy of a plant factory. Furthermore, the ability to automatically perform such environmental controls is essential in a plant factory, leading to research into technologies for precise control within the plant factory [2] and technologies that allow real-time monitoring of the plant factory from remote locations [3].

In contrast, soybean sprouts have a short cultivation period and do not require light or $CO_2$ as they do not photosynthesize. They can be cultivated by merely controlling water temperature and ambient temperature. Therefore, anyone can cultivate soybean sprouts with relatively low technical costs, making it difficult to find publicly available research materials on automated cultivation systems.

Moreover, the anatomy of plants grown in a plant factory is generally more complex than that of soybean sprouts. While soybean sprouts consist only of seeds and roots, most crops grown in plant factories have roots, stems, and leaves, and in some cases, flowers and fruits. Therefore, the nutrient ion control technology required for fertilization in a plant factory is generally of a very high standard [4]. Interactions between various ions can distort the voltage value of the Ion Selective

Electrode by up to 40%, leading to research into machine learning technology for correction [5, 6]. Additionally, technologies for analyzing the chemical complex in the nutrient solution [7] and complex system technologies analyzing the interaction between the nutrient solution and the plant system are also being researched [8].

On the other hand, soybean sprouts grow by consuming the nutrients contained in the beans before germination, and these nutrients are abundant enough to withstand seven days of cultivation without the need for external nutrient supplementation. Therefore, there has been little research into the composition of the water used for cultivation or the interaction between water and plants in the case of soybean sprouts.

*D. Automated Quality Control Technology Required*

The ease of mass production has led to a situation where the level of quality control becomes a competitive edge for the product. Not only is it important to use good raw materials to maintain a high level of quality uniformly, but it is also crucial to meticulously review the production and packaging processes to produce better products. This is the direction of competition that the modern soybean sprouts industry should take. Since the cultivation period of soybean sprouts is very short, only 7 days, it is easy to monitor the entire process of each cultivation cycle, and the results of cultivation can be quantitatively analyzed to quickly repeat the process of improving the cultivation environment.

However, it is physically impossible to measure each of the millions of soybean sprouts individually, and even sampling surveys take a considerable amount of time. It took us about six hours to measure the weight of 500g of soybean sprouts and the head length, body length, body thickness, and tail length for the creation of the dataset. It is nearly impossible to measure these on an industrial site every day. Therefore, the development of technology to automatically measure the results of soybean sprout cultivation is absolutely necessary.

*E. The Absence of Soybean Sprouts Datasets*

There were no data-driven analyses or available datasets related to the cultivation of soybean sprouts. Various studies have been conducted on soybeans themselves. For instance, Yang et al. [9] proposed a dataset for semantic segmentation on soybean pods and suggested an analysis technique using this dataset. Wang et al. [10] proposed a cultivar recognition technology through pattern analysis of soybean leaf image data. Abade et al. [11] proposed an AI model that distinguishes nematodes that infect soybeans, and Griera et al. [12] proposed a deep learning technology to estimate soybean production. Such various cases of research on the cultivation process of soybeans, a major crop, are being reported worldwide. These cases also provide important resources by disclosing the datasets used in the research, allowing other researchers to refer to them and train AI models.

However, there are no sophisticated datasets related to soybean sprouts, leaving industry practitioners without resources to reference for research. It makes AI based approach on soybean sprout production much more difficult.

*F. Necessity of New Dataset for Advanced Analysis*

Considering the substantial demand for soybean sprouts in Korea and the global consumption of mung bean sprouts, which can be cultivated similarly, there is an undeniable advantage to applying advanced analysis to soybean sprout images. Therefore, the creation of a sophisticated image dataset using soybean sprouts is crucial. We have constructed a fundamental dataset by photographing soybean sprouts on a large scale and through manual labeling. We then generated a significant number of augmented images and mask labels through the combination of each image.

Our dataset, CongNaMul, supports the classification task that can distinguish whether there are diseases or spots, or whether the stem is broken for the quality control of produced soybean sprouts. It also encourages precise analysis using semantic segmentation and image decomposition and provides a dataset for automated physical feature measurement. This dataset can contribute to the development of technology for automated quality control at the soybean sprout factory. This paper provides a description of this dataset.

## II. METHOD

*A. Soybean Sprout Sample Preperation*

Soybean sprouts from Pulmuone, CJ, and Haetrac were each purchased in quantities of 4kg for image capture. Any samples which had been out of the bag for more than 4 hours were discarded. While it is difficult to precisely calculate the time elapsed from harvest to sample photography, it is estimated that the sprouts photographed were harvested within approximately 48 hours, considering that products harvested within 2 days are displayed in the B2C Market due to the distribution process of soybean sprouts cold chain in Korea. We photographed the samples directly without any washing or cleaning.

*B. Raw Image and Physical Features Acquisition*

To maintain a uniform scale in all images, we standardized the shooting conditions and camera. All photos were taken using the default camera app on the Samsung Galaxy S22, resulting in 3,024x3,024 JPEG format images. All photos were taken with the bean sprouts placed on the ground, with the camera lens surface fixed at a height of 30.5cm from the ground.

Photos were taken against three types of backgrounds, as can be seen in Figure 1. (a). We decided to take photos against green and white checkered backgrounds in addition to a clear background, anticipating that it would be difficult to create a robust model with data from a clear background alone. Green is a color applied to many industrial conveyor belts in Korea and contrasts with the color of the bean sprouts, making them stand out clearly. White is a color commonly applied to food conveyor belts in Korea, and since the body of the bean sprouts is also white, it is relatively difficult to distinguish compared to green.

The green and white backgrounds have a distinct grid of a different color. Each grid is arranged at 5mm intervals in each cell. This grid pattern serves as a kind of interference factor, making AI learning a little more difficult and expected to help it operate more robustly even in real backgrounds with foreign substances. In addition, if you are developing a length measurement technology based on simple pixel analysis rather than a neural network-based model, the pattern of the grid arranged at a constant interval will help develop the measurement algorithm.

Each bean sprout was re-photographed against three different backgrounds. An example of a single sample image can be seen in Figure 1. (b). In addition, for more complex tasks, multiple sample images were taken with five bean sprouts randomly placed in one shot, which can be seen in Figure 1. (c). After taking the photo, the physical features of the bean sprout were measured. The weight was measured using an electronic scale, and then the researcher meticulously measured the length of each part using a vernier caliper.

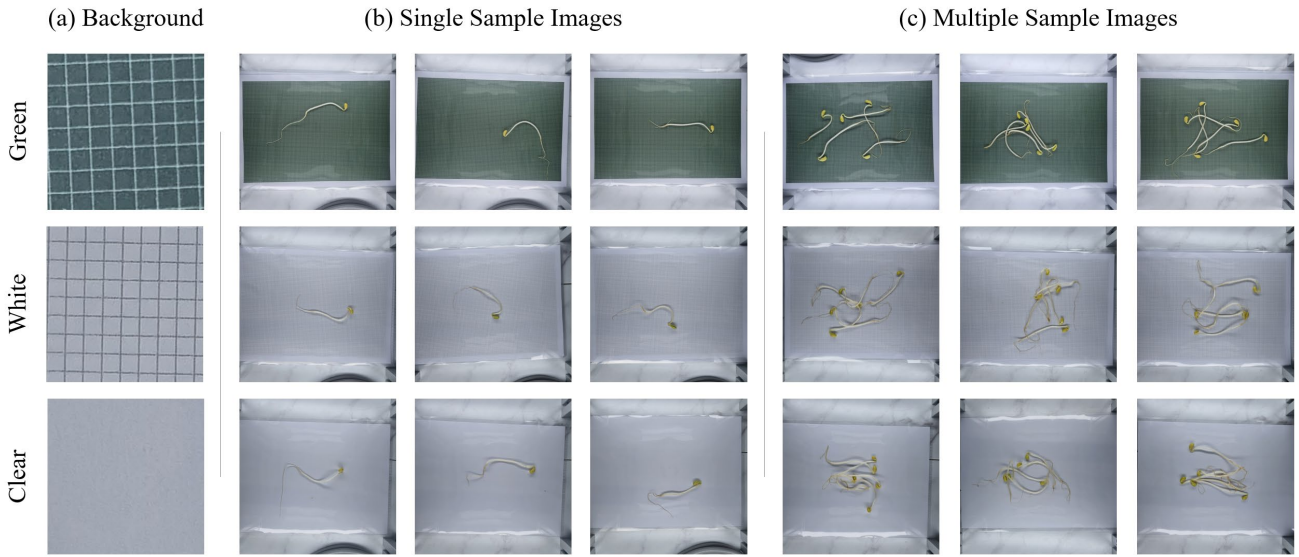

Fig. 1. Raw image samples.

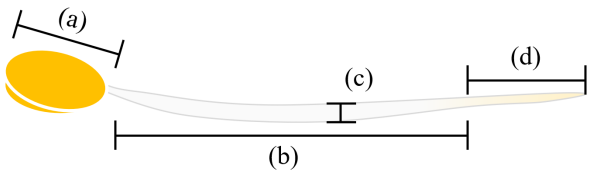

Fig. 2. (a) head length, (b) body length, (c) body thickness, (d) tail length

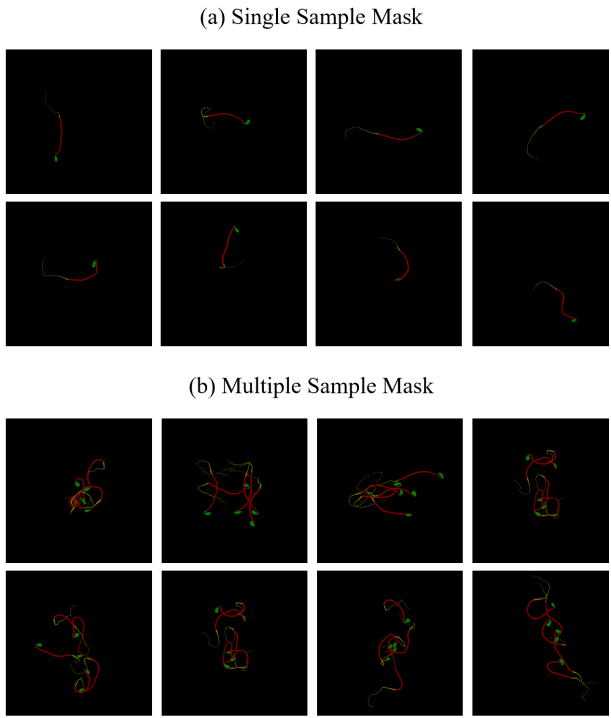

Fig. 3. Mask images for semantic segmentation task

Length measurements were performed targeting the head length, body length, body thickness, and tail length, and the criteria for dividing each part can be seen in Figure 2.

A total of 205 bean sprouts were photographed, and of these, 45 bean sprouts were missing weight measurements. Since each of the 205 bean sprouts was photographed on three separate backgrounds, a total of 615 single bean sprout photos were collected. For multiple sample images, five bean sprouts were collected at a time, randomly scattered each time, and photographed 10 times on three types of backgrounds. Therefore, a total of 3,690 multi-sample images were collected. To improve the quality of the dataset, images with shaky focus were removed from the collected photos. As a result, a total of 604 single samples were collected, and a total of 1,030 multiple samples were collected, and physical features were collected for all these photos.

Subsequently, photos were taken in the same way for the production of the classification dataset, but physical features were not measured at this time. Detailed explanations will be provided in the classification dataset section.

*C. Semantic Segmentation Dataset Production*

We utilized the Labelme [13] to segment the captured images. Using the software, we distinguished the objects in the photo into three classes (head, body, tail), and the remaining areas were automatically labeled as _background by Labelme. The labeling results were output in JSON format, defining polygons and their coordinates for each class.

While the JSON format can efficiently store the coordinates of each point and reduce capacity, palette mode PNG files can be directly input as labels for semantic segmentation. Considering the convenience of training, we converted all images to palette mode PNG files. These PNG files only contain four unsigned int values: 0, 1, 2, 3. You can download and use the data that suits your situation from either the JSON format mask or the PNG format mask. If you are conducting training with TensorFlow, it is convenient to input the PNG file in Numpy array format and use a loss function such as sparse_categorical_crossentropy(), which substitutes for the one-hot conversion of data.

The labeling results for segmentation are represented in Figure 3.

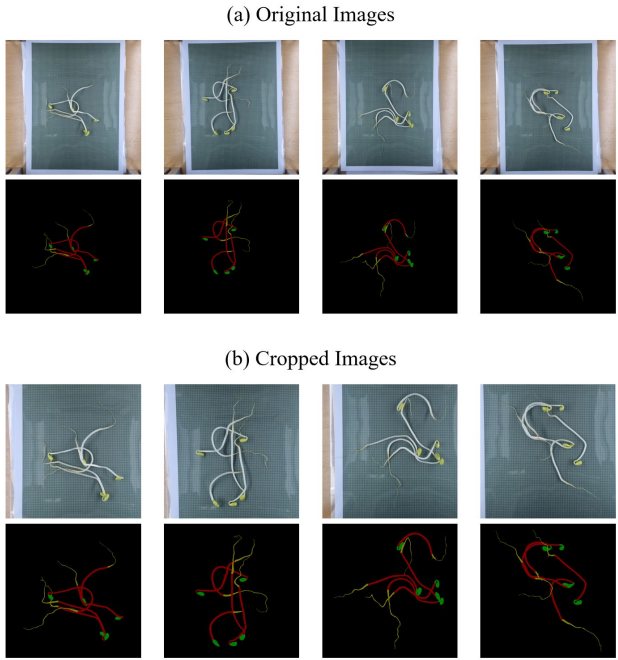

Fig. 4. Image Cropping results.

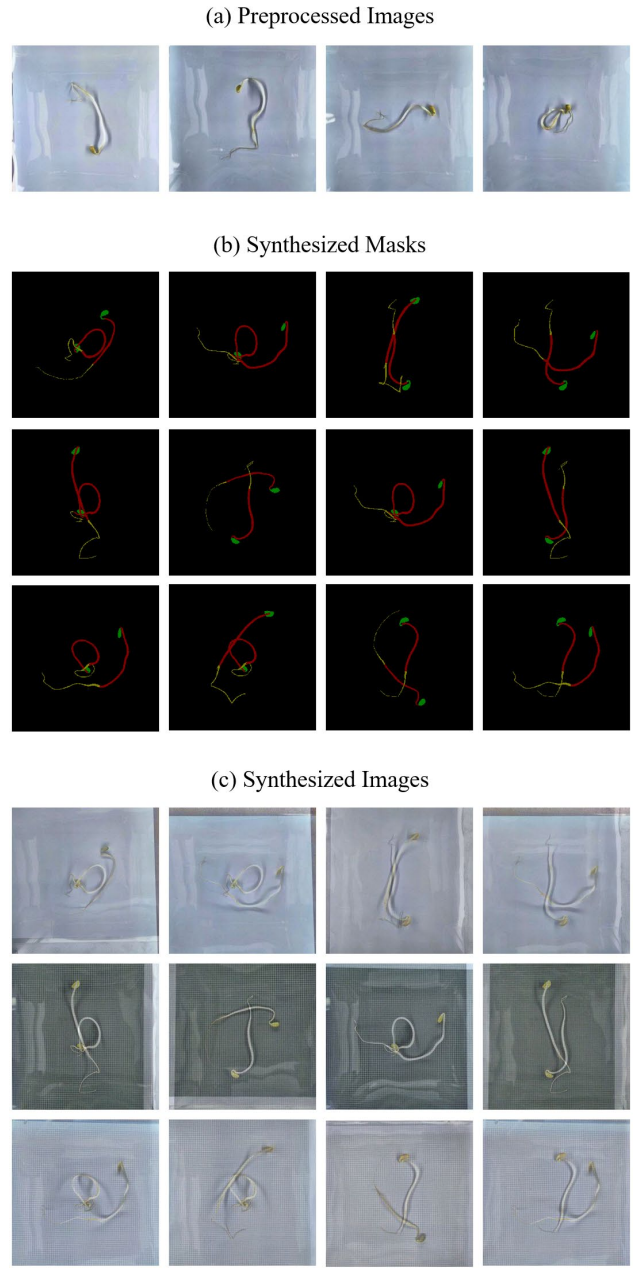

Fig. 5. Image decomposition dataset overview.

*D. Image Cropping with Segmentation Mask Analysis*

The human-labeled segmentation mask contains precise information about the location and size of the soybean sprouts. Therefore, by analyzing the pixel distribution of the labeled mask, it is possible to crop the image to the smallest size while ensuring that the entire appearance of the soybean sprout is fully represented in the photo.

We analyzed the mask PNG file of single sample images and discovered that the image of the longest soybean sprout occupies 2,016px in one direction. Therefore we cropped the 3024x3024px size image to convert it into a 2,016x2,016px size image. We also analyzed the mask of each photo to calculate the starting and ending points of cropping so that the soybean sprout is always located in the center of the photo, and cropped the image at the same coordinates. Multiple sample images were cropped into 2,675x2,675px images with same procedure.

This was done to eliminate areas unnecessary for training, thereby enhancing the efficiency of training and optimizing the VRAM capacity consumed during training. An example of the cropping result is presented in Figure 4. If you wish for more robust model training, you can still download the uncropped original images.

*E. Image Decomposition Dataset Production*

In industrial situations, there are hardly any processes that check soybean sprouts individually. In most automated factories, hundreds of thousands of soybean sprouts are transported at once through conveyor belts or other automated devices. Even in production factories where automation has not been introduced, it is common to harvest, wash, or bag soybean sprouts in hundreds of grams at a time, and there is no process to check them one by one. Therefore, it is impossible to photograph each soybean sprout one by one and input it into an AI model in the actual production process. Most of the photos that can be collected in the factory are in the form where multiple soybean sprouts are photographed in one photo at the same time. In fact, most of them are even tangled or overlapped with each other.

Therefore, it is necessary to develop a model that takes a single photo with multiple tangled soybean sprouts as input and decomposes it into multiple photos with a single soybean sprout. To develop an artificial intelligence model that can perform such tasks, we have created an Image Decomposition Dataset.

We processed the images by combining two images to create a new single image. The synthesis result can be seen in Figure 5.

Firstly, preprocessing was performed to easily combine photos taken in slightly different environments. The contrast limited adaptive histogram equalization [14] technique was applied to all images to equalize the distribution of pixel values. The preprocessing result is in Figure 5. (a). As a result of preprocessing, bright backgrounds and areas of light reflection have become somewhat darker.

Secondly, mask synthesis was performed. Two masks were loaded as Numpy arrays of size (2016, 2016). Since the masks were made as palette type PNG, there is no separate color channel. A new array of size (2016, 2016) initialized with all values as 0 was created, and the values of each coordinate of the two mask arrays were analyzed to create a new mask with two soybean sprouts drawn. The pixel comparison algorithm is very simple. By analyzing the pixel values of the two arrays, the larger value of the two was recorded in the new array. However, there is a limitation that class 3 always covers class 1, 2, and class 2 covers class 1. The creation result is in Figure 5. (b).

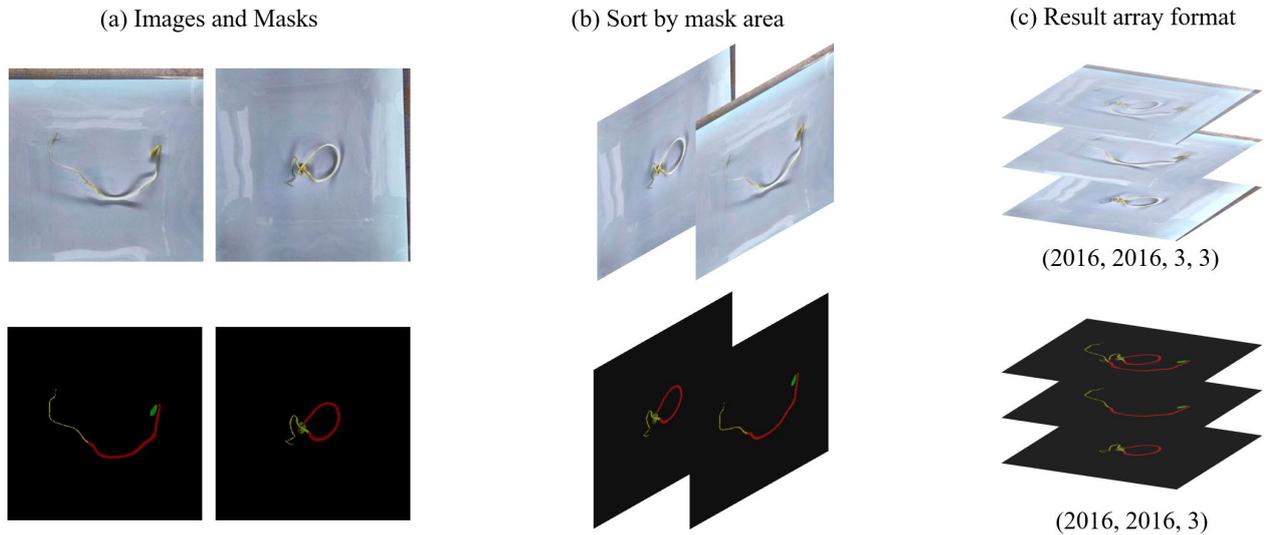

Fig. 6. Array building strategies. (a) raw images and masks to merge, (b) sort the data by non-zero area of mask, (c) the array format and dimensions of training-ready data for images and masks.

Thirdly, image synthesis was performed. Generally, pixel values of image files are represented as uint8 type data, but we converted this data to float32. A new image was created by calculating the average of the pixel values of the two images. Figure 5. (c) shows the result of this work.

In the process of selecting the two photos to be synthesized, only images taken from the same background pattern were selected to maintain the advantage of providing three distinct types of backgrounds for robust learning. 200 clear background images, 204 green checkered background images, and 200 white checkered background images were used. Two photos were selected at once and synthesized without duplication, and this process exactly followed the combination $_nC_2$. The number of synthesized images produced is 19,900 for white and clear backgrounds, and 20,706 for green backgrounds. Therefore, the total number of this dataset is 60,506.

Since the result of the combination operation increases in a nonlinear scale, the size of this dataset is too big for an open storage service. So we provide this dataset as source image files and generator source codes, ot raw files. The generator code produces separately organized .npy file format Numpy binary data.

The process of decomposing multiple data from a single data typically involves elements to be decomposed having distinct features. For example, in the process of decomposing violin and horn sounds from orchestra sounds, the violin and horn sounds to be separated physically have different features called "timbre". On the other hand, in the case of bean sprouts, it is technically a bit tricky because you have to distinguish between two bean sprouts with completely the same shape and features.

Among them, what makes end-to-end learning most difficult is in what order the merged image received as input is separated. For example, when an image created by synthesizing bean sprout A picture and bean sprout B picture is put into the model, when the model generates a decomposed image in the order of (A, B) and when it outputs an image in the order of (B, A), which one is the correct answer? In essence, both are correct, but designing a loss function that considers both cases as correct is cumbersome. If you divide the bean sprout picture into 2 pieces, you can perform 2 min() operations because there are 2 possible order pairs. If the bean sprout picture becomes 3, 6 operations are needed. Thus, a loss function that works regardless of the order of the decomposed images output by the model is expected to operate with $O(n!)$ time complexity.

To alleviate such learning inconvenience, we designed the structure of this data with a resonable standard, which is represented in Figure 6. (c). The binary file containing images is in the form of (width, height, RGB, 3 images), and the binary file containing masks is in the format of (width, height, 3 masks). We analyzed the masks used in the synthesis to calculate the area of the bean sprouts marked on each mask, and sorted the images and masks in order of increasing mask area. And the image with a smaller mask area was placed in the front channel of the result array, and the image with a larger mask area was placed in the back channel of the result array. And the synthesized image was attached to the last channel.

Figure 6. shows the case of creating a dataset by synthesizing 2 images. The last channel indices 0 and 1 of this result array correspond to the ground truth of the decomposed image that the model needs to infer, and the index -1 (or index 2 in this case) value is the data given as the input of the model. Researchers can use this binary file to always perform end-to-end learning so that the model divides multiple decomposed images from smaller to larger areas, which is also a way to solve the problem of the operation time of the loss function increasing by $O(n!)$.

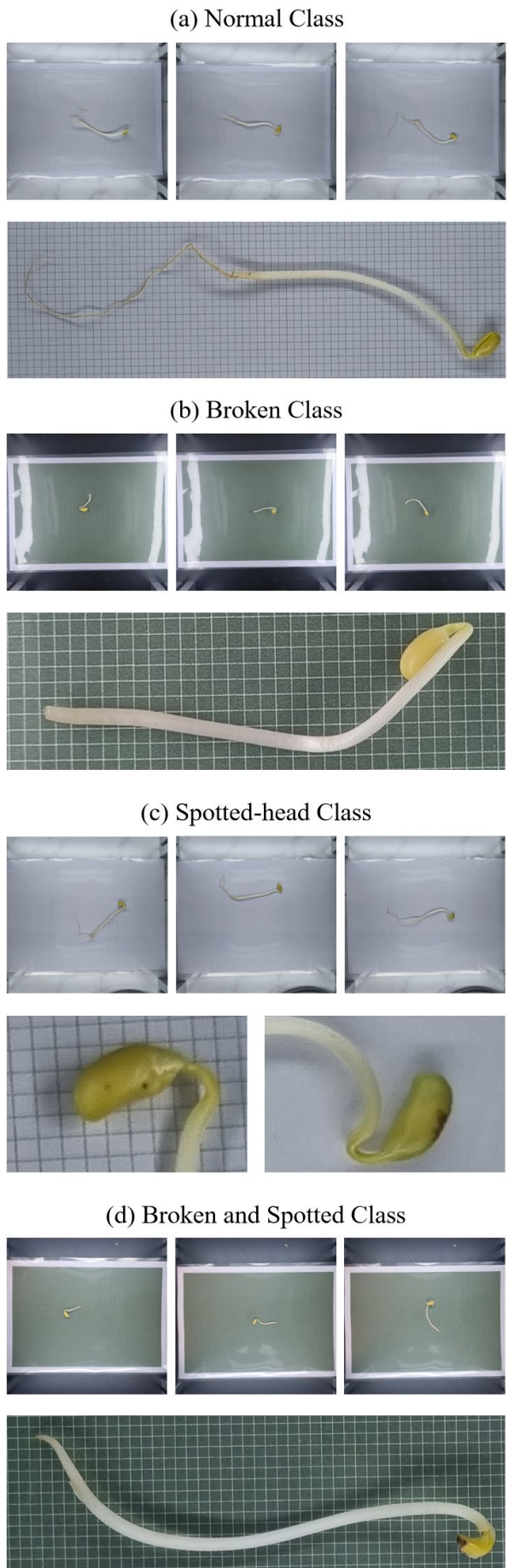

Fig. 7. Classification dataset overview.

## F. Classification Dataset Production

We performed the task of directly classifying 615 single bean sprout photos into four categories. For the increase in dataset volume and balance between classes, we additionally photographed and classified 1,385 photos. Each class consists of 500 photos, and the classification results are shown in Figure 7. All photos are 3024x3024px JPEG files and no preprocessing was performed.

The first class, "Normal", is a bean sprout photo of the highest quality without any damage.

The second class, Broken, is a bean sprout individual with a broken body (hypocotyl). Broken class samples mainly occur during the stages of harvesting, washing, and packaging the grown bean sprouts. Industrially, it is not fatal if the tail (root) is broken, and some restaurants even remove the tail of the bean sprouts before using them in cooking. On the other hand, if the body is broken, it is not a welcome state industrially and is an example of low quality. Reflecting such industrial standards, we classified only samples with damaged body parts as the broken class.

The third class is spotted-head. This is an individual with spots on the head (cotyledon, leaf) of the bean sprout. Good quality bean sprouts have a smooth and glossy yellow head. However, some individuals may have black spots due to various physiological influences such as low quality of raw beans, infection during cultivation, and failure to control environmental factors. Such spots are the most fatal defect factor because they stand out even after cooking.

The fourth class includes bean sprouts that fall under both broken body and spotted head.

## G. For Physical Features Prediction Task

Inferring the size of each part of an object solely from photos is a demand that would exist in any industry. In the horticultural industry, measuring the physical features of harvested crops greatly aids in achieving optimal cultivation conditions. In the case of the bean sprout cultivation industry, with a very short growth cycle of 7 days, frequent optimization is possible, making the task of measuring individual information even more valuable.

To aid in the creation of a dataset that performs this, we provide length information for all bean sprout individuals included in the Semantic Segmentation Dataset and Image Decomposition Dataset. This information is provided in json format, and researchers can easily check the physical characteristic values by searching for individual file names in the json data. The format of the json file is as follows.

```
{
  <filename>: {
    "length_head": <Number>,
    "thickness_body": <Number>,
    "length_body": <Number>,
    "length_tail": <Number>,
    "weight": <Number>
  },
  ...
}
```

All length information is expressed in millimeters (mm) as a float, and weight is in milligrams (mg). However, for samples where weight measurement was omitted during the raw data measurement stage, the "weight" value is marked as -1.

| | Semantic Segmentation Dataset | | Decomposition Dataset | Classification Dataset |
|---|---|---|---|---|
| **Type** | Single Sample | Multiple Samples | Two samples | Single Sample |
| **Number** | 604 | 1,030 | 60,506 | 2,000 |
| **Size 1** | 3,024 x 3,024 (2,016 x 2,016 cropped) | 3,024 x 3,024 (2,675 x 2,675 cropped) | Image: (Xdim, Ydim, 3, 3) Mask: (Xdim, Ydim, 3) Free Resizing Available | 3,024 x 3,024 |
| **Size 2** | 1,024 x 1,024 (cropped) | 1,359 x 1,359 (cropped) | | 1,024 x 1,024 |
| **Size 3** | 512 x 512 (cropped) | 679 x 679 (cropped) | | 512 x 512 |
| **Size 4** | 256 x 256 (cropped) | 340 x 340 (cropped) | | 256 x 256 |
| **Format** | Image: JPEG MASK: PNG (palette) | | Generator Source Code (.py script) | JPEG |
| **Physical Features** | Yes | Yes | Yes | No |
| **Classes** | 4 | | - | 4 |

Table 2. Dataset statistics overview.

For the multiple samples in the Semantic Segmentation dataset, physical feature information for the five bean sprouts included in the photo is provided all at once, and it is not provided which bean sprout object depicted in the photo corresponds to which value in the json.

### III. DATASET STATISTICS

The statistics of the entire dataset are summarized in Table 1. The data shown in this table are without any separate augmentation, so a researcher can apply various augmentation methods to establish a larger dataset for training. For a simple and basic example, simple vertical and horizontal flips can inflate the data by four times, and adding a few more 90-degree rotations can easily increase the number of data.

The Semantic Segmentation dataset is composed of 615 single sample cases and 1,030 multiple sample cases. If the purpose is to perform simple segmentation tasks, it is recommended to mix all of them and perform training on 1,675 cases. If the purpose is to measure the length of a single object, it is recommended to take only the Single sample dataset and use it for augmentation.

The Decomposition dataset was created only for cropped images. The dataset's capacity was too large, so data cropping was done to reduce the capacity a bit. Using the dataset production code we provide to directly create images of the size needed by the researcher is also a good option.

The Classification dataset simply provides four folders with 500 images each. Each folder represents the class of the soybean sprout.

### IV. APPLICATION STRATEGIES

To measure the size of each part of a bean sprout, it is first necessary to distinguish which area in the picture corresponds to which part. Therefore, training a preprocessing model that performs semantic segmentation may be necessary for such a task. The Semantic Segmentation Dataset can be used for training such a model. Also, because physical features are provided, it can be used for training a model that learns the length of bean sprouts in an end-to-end manner, or a model that predicts the physical characteristics of bean sprouts by receiving the pixel distribution of an AI-predicted mask.

The Decomposition Dataset is essential for developing advanced analysis models of bean sprouts that can be used in real industrial environments. It is impossible to take pictures of individual bean sprouts and collect them on a large scale in a bean sprout factory. Therefore, a process that decomposes a single picture, in which a large number of bean sprout entities are photographed together, into multiple pictures of individual bean sprouts must be performed. The Decomposition Dataset can contribute to the training of such a model.

Finally, the Classification Dataset can be used to develop a model that investigates the quality of bean sprouts simply using a classical CNN-based classifier. With simple augmentation, it is expected that an efficient-sized AI that can be mounted on a mobile device can be trained.

### V. CONCLUSION

The CongNaMul dataset provides several types of datasets that can facilitate the development of various AI technologies needed in the production field of bean sprouts. By utilizing this dataset, various models that can be used for quality control of bean sprouts can be trained.

However, the size of the dataset could be a limitation. Although the Decomposition dataset is sufficiently large, the single sample segmentation data is small in size and requires augmentation during the training process. We hope for the development of a larger dataset that overcomes these limitations in the future.

But the high accuracy of the data is a significant advantage of this dataset. Not only does it contain conventional human-labeled segmentation masks, but it also includes physical features measured directly by human researchers. Models trained by researchers with reliable data will likewise become reliable AI technologies.

ACKNOWLEDGMENT

We would like to express our deepest gratitude to Seung Yeop Jang, Tae Dong Eom, Changyul Lee, Minwoo Lee, Yeongbeom Kwon, and Junsan Kim for their invaluable assistance in the semantic segmentation labeling of soybean sprout images. Their contributions were crucial in the development of the 'CongNaMul' dataset.

SUPPLEMENT DATA

Dataset and overlapping code are available at https://bhban.kr/data.